\documentclass{article}

\usepackage{neurips_data_2023}
\usepackage[pdftex]{graphicx}
\usepackage[colorlinks=true,linkcolor=black,anchorcolor=black,citecolor=black,filecolor=black,menucolor=black,runcolor=black,urlcolor=black]{hyperref}
\usepackage{wrapfig}
\usepackage{amsfonts}
\usepackage{amsmath}

\usepackage[utf8]{inputenc} 
\usepackage[T1]{fontenc}    
\usepackage{hyperref}       
\usepackage{url}            
\usepackage{booktabs}       
\usepackage{amsfonts}       
\usepackage{nicefrac}       
\usepackage{microtype}      
\usepackage{xcolor}         

\title{LiveChat: Video Comment Generation from Audio-Visual Multimodal Contexts
}

\author{%
  Julien Lalanne\\
  National Institute of Informatics\\
  Tokyo, Japan \\
  \texttt{ julien.lalanne.travail@gmail.com} \\
  \And
  Raphael Bournet \\
  National Institute of Informatics\\
  Tokyo, Japan \\
  \texttt{raphael.bournet@u-psud.fr} \\
  \AND
  Yi Yu \thanks{Yi Yu is the corresponding author of this work.}\\
  National Institute of Informatics \\
  Tokyo, Japan \\
  \texttt{yiyu@nii.ac.jp} \\
}

\begin{document}

\maketitle

\begin{abstract}
         Live commenting on video, a popular feature of live streaming platforms, enables viewers to engage with the content and share their comments, reactions, opinions, or questions with the streamer or other viewers while watching the video or live stream. It presents a challenging testbed for AI agents, which involves the simultaneous understanding of audio-visual multimodal contexts from live streams and the ability to interact with human viewers through dialogue. As existing live streaming-based comments datasets contain limited categories and lack a diversity, we create a large-scale audio-visual multimodal dialogue dataset to facilitate the development of live commenting technologies. The data is collected from Twitch, with 11 different categories and 575 streamers for a total of 438 hours of video and 3.2 million comments. Moreover, we propose a novel multimodal generation model capable of generating live comments that align with the temporal and spatial events within the video, as well as with the ongoing multimodal dialogue context. Our initial results have demonstrated the effectiveness of the proposed model, providing a robust foundation for further research and practical applications in the field of live video interaction.
\end{abstract}

\section{Introduction}
Live commenting on videos has become a very popular feature in live video platforms such as Twitch, YouTube, Facebook, and Instagram. Live video comment generation is able to provide viewers with a more interactive and dynamic experience during live video content, fostering engagement and encouraging discussion among viewers. Live video comments generated from users tend to be divergent yet informative, which are different from video captions where objective descriptions of the visual semantic contents are provided. Additionally, it is challenging to ensure that the generated comments are relevant to the ongoing video content or discussion.

\begin{figure}
  \centering
  \includegraphics[scale=0.2]{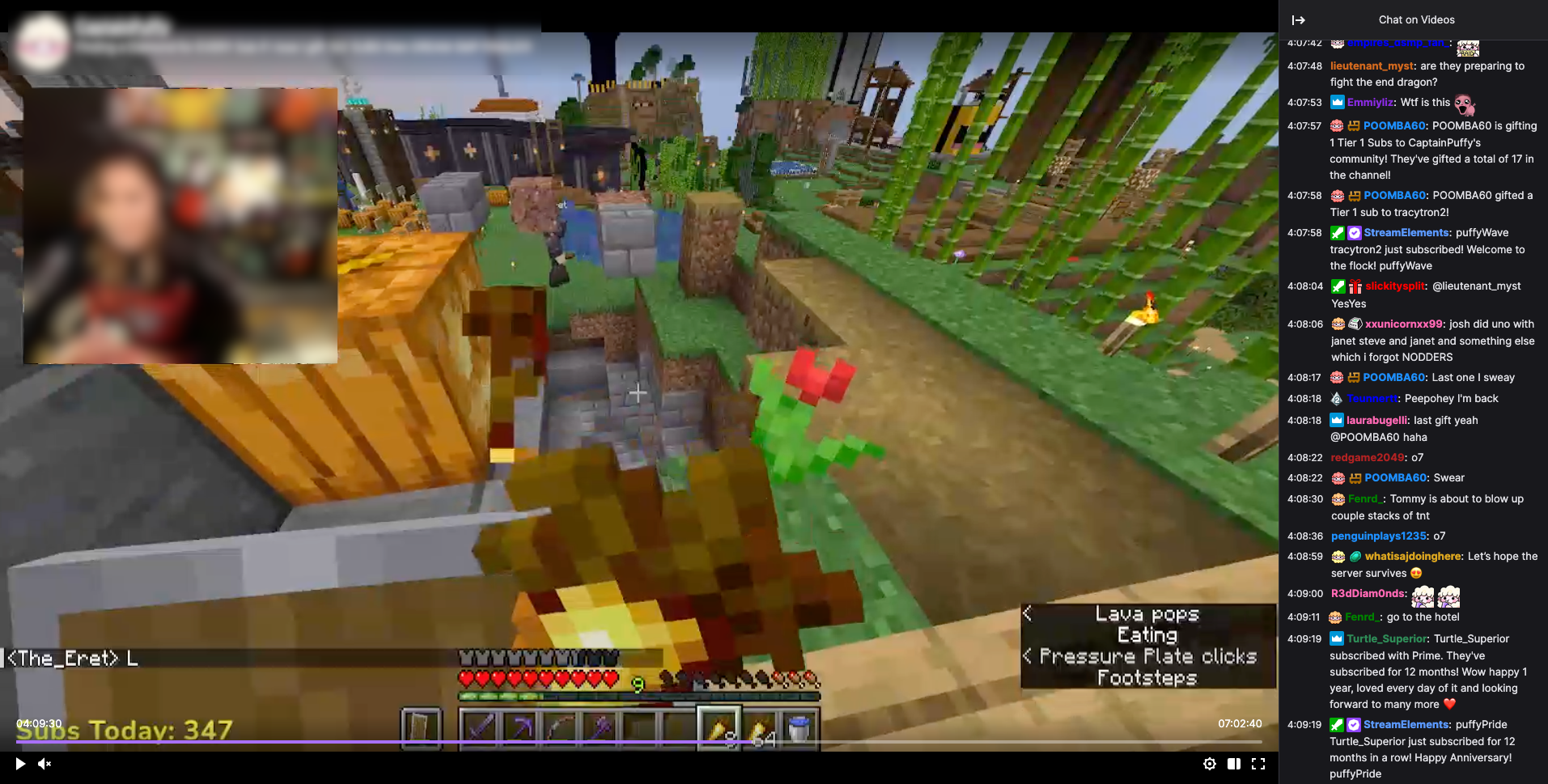}
  \caption{A live video example viewed on the Twitch website.}
  \label{fig:1}
\end{figure}

Among existing live video comment datasets, Livebot[7][16], VideoIC[13], were collected from the same Chinese video platform Bilibili\footnote{\url{https://www.bilibili.com/}}, while Twitch-FIFA Dataset[8]  was collected from soccer gaming videos in Twitch\footnote{\url{https://www.twitch.tv/}} and VideoLC[1] from the Chinese movies platform qq\footnote{\url{https://v.qq.com/}}. As these datasets contain the limited categories and lack a diversity, more large-scale audio-visual multimodal video comment datasets are indeed necessary to complement the development of live commenting technologies. Twitch is now not only used for gaming and diversified its content, which includes categories like music, art, talk shows, and more, attracting a variety of content creators and viewers. We create a large-scale, diverse audio-visual multimodal video comment dataset with popular categories from Twitch, which can be very helpful to advance research in this field. Our LiveChat dataset \footnote{\url{ https://github.com/yy1lab/LiveChat}} contains 11 different categories and 575 streamers for a total of 438 hours of video and 3.2 million comments. Each video is conditioned by the live comments thus allowing more complex interdependencies. An example of such live comments can be seen in Figure \ref{fig:1} in which a content creator is playing the video game Minecraft and the people watching her are commenting on what she is doing or asking her questions. It is obvious that these video-based contexts contain important multimodal dependencies as the comments are either a consequence of the action in the video or giving additional information about the video itself.

In addition, a multimodal generation model is suggested to generate live comments from the video which are relevant temporal and spatial event language, as well as relevant to the multimodal dialogue context. The model uses an attention mechanism in order to attend each modality and generate an appropriate comment. Our initial results have demonstrated the effectiveness of the proposed model.

\section{Related Work}
This work builds upon existing research in the area of Live Comment Generation.

The field of \textbf{Live Comment Generation} has been introduced in [7] where the authors proposed a "bullet screen" benchmark dataset extracted from Bilibili. They also established two baselines using Long Short-Term Memory (LSTM)[2,3,5,6] and Transformer [12] architectures.

Subsequent works have addressed various challenges and improvements to this task. In [15], the authors highlighted some shortcomings with this first baseline, leading to further investigations into optimizing the task's results. Following this works, [11,13,16,17,18] focused on using better metrics and generation strategies as well as enhancing the dataset itself. Notably, [1] aimed to generate more contextually relevant comments by using external knowledge [14].

The majority of the works mentioned have relied on Danmu data. To our knowledge, the only work directly addressing the Live Streaming data is [8]. This study focused on video-based dialogue generation, creating the Twitch-FIFA dataset from soccer videos on the Twitch platform. Their research further explored both discriminative and generative models using LSTM and attention mechanisms to approximate the joint probability distribution.

This current study stands apart in the field by focusing on the quality of the dataset and the innovative use of transformer architectures. While previous works focused on improving metrics and generation strategies, our research emphasizes the creation of a large-scale, high-quality multimodal live-streaming dataset, addressing an existing gap in the area. Furthermore, the application of transformer models in our approach represents a novel method to generate live comments. Together, these contributions set the stage for a new direction in live commenting technologies, offering substantial promise for future research and practical applications.

\section{Problem Statement}

Live Comment Generation in the context of videos involves generating relevant and coherent comments in real-time based on multiple input sources, including the video content, audio content, and existing comments from viewers. Let \( \mathcal{V} \) represent the set of all available videos, \( \mathcal{A} \) be the set of audio features, and \( \mathcal{C} \) be the set of existing comments. The goal is to generate a new comment \( r \) that is contextually appropriate and informative.\\

Each video \( v \in \mathcal{V} \) can be represented as a sequence of frames \( \{f_1, f_2, ..., f_T\} \), where \( T \) is the total number of frames in the video. To capture the video context, we extract visual features \( \{v_1, v_2, ..., v_T\} \) using a ResNet model, where \( v_i \) is a \( d_v \)-dimensional vector representing the visual information from the \( i \)-th frame.

In addition to visual information, each video \( v \) also contains audio content. We convert the audio features into text representations \( \{a\} = \{a_1, \dots, a_{N_a}\} \) using a speech-to-text model. \( a \) is a \( d_a \)-dimensional vector capturing the text information contained in the audio of the video.

The existing comments \( C \) for a given video \( v \) form a set of \( N \) comments, \( C = \{c_1, c_2, ..., c_N\} \), where \( c_i \) is the \( i \)-th comment. Each comment \( c_i \) is associated with a timestamp \( t_i \) indicating when the comment was posted during the video playback.

The goal of Live Comment Generation is to generate a new comment \( r \) that is contextually relevant to the video and audio content, as well as coherent with the existing comments. This can be formulated as finding the most probable comment given the video and audio contexts and the set of existing comments:

\begin{equation}
r = \arg\max_{c'} P(c' | v, a, c)
\end{equation}

where \( P(c' | v, a, c) \) represents the conditional probability of generating comment \( c' \) given the video context \( v \), audio context \( a \), and existing comments \( c \).\\

Generating live comments in a video context poses several challenges:

\begin{itemize}
  \item \textbf{Context Fusion}: Effectively combining information from video, audio, and existing comments to generate contextually appropriate comments.
  \item \textbf{Coherence with Existing Comments}: Ensuring that the generated comments are coherent with the sentiments and themes expressed in existing comments.
  \item \textbf{Diversity and Creativity}: Striving to generate diverse and creative comments rather than repetitive or generic responses.
\end{itemize}

Addressing these challenges will lead to the development of more sophisticated and effective Live Comment Generation systems.
\newpage
\section{LiveChat Dataset}

\subsection{Data Collection and Preprocessing}
The content that we propose in our dataset is extracted from the Twitch website. While Twitch's main public is oriented around video games content, we try to extend our dataset to a wide variety of the content available on the plateform. The different categories and their representation can be seen in Figure 2. Half of the categories correspond to popular video games on the website, where the other half are related to a wide range of subjects such as music or arts which are categories that may invite the viewers to write longer and more coherent comments than the one we can usually see in video games category. We believe that expanding the range of categories will make the dataset applicable to a broader spectrum of tasks.
\vspace{0.5cm}

\begin{wrapfigure}{R}{0.5\textwidth}
    \centering
    \begin{tabular}{cc}
    \hline
    Category & Number of extract\\
    \hline
    Fortnite & 292 \\
    League of Legends & 230 \\
    World of Warcraft & 258 \\
    Special Events & 225 \\
    Overwatch 2 & 240 \\
    Minecraft & 243\\
    Music & 162\\
    CS: Go & 213\\
    Art & 225\\
    Just Chatting & 267\\
    Valorant & 273\\
    \hline
\end{tabular}
    \caption{Repartition of the videos across the categories }
    \label{tab:3}
\end{wrapfigure}

To collect the videos, we use the following process:
using the Twitch API\footnote{\url{https://dev.twitch.tv/docs/api/}}, we first find the top viewed categories and manually choose 11 of them to ensure the previously explained diversity. Then, we retrieve in each category the top viewed live streams of the week for four consecutive weeks and filter them to get at most two videos per week of the same streamer. Using the open-source tool TwitchDownloaderCLI\footnote{\url{https://github.com/lay295/TwitchDownloader/tree/master/TwitchDownloaderCLI}}, we download the entierity of the comments from these videos and finally use these comments to find the 30 minutes with the best comment density in each video and download these parts with the same tool.\\

We then perform several transformations on these raw videos to enhance the dataset's usability, ensure anonymity, and reduce the storage requirements. Each video is first reduced from 1080p60 to 720p5 to conserve space, while maintaining adequate quality and frame count for various tasks. Each frame is then encoded into a $2048$-dimensional vector through a ResNet50 pretrained on ImageNet, to anonymize the videos and facilitate computation. Since most audio information in these videos corresponds to the streamer's speech, we employ a pretrained Speech2Text model to transcribe the content. Additionally, we remove usernames from the data to preserve privacy.

Finally, each video is divided into 30-second clips and sampled at 1 frame per second for our specific task. Each clip consists of a 20-second context window and a 10-second response window.

The processed dataset will be publicly available on GitHub, and the raw data can be provided upon request, allowing researchers and practitioners to explore and utilize the data according to their specific needs and interests.

\subsection{Dataset Analysis}
\label{dataset_analysis}

The dataset we have constructed is compared to four other key datasets in the domain, as outlined in Table \ref{tab:1}. Three of these are considered reputable and are constructed with Danmu comments, but are in Chinese. The fourth dataset, although also extracted from Twitch, demonstrates limitations in several key areas.

Our dataset offers a broader collection of videos, representing a wide array of content and categories within the English-speaking context. With a diverse collection of comments, it captures a wide range of English language user interactions and reactions. The total duration of our dataset and our emphasis on segments with the highest comment density ensures that it is focused on the most engaging parts of the videos, enhancing its relevance for studying live English-language interactions.

Unlike the three Chinese datasets, our dataset is in English, filling a critical gap in the existing landscape. Our comparison with another dataset derived from Twitch, but of lesser quality, accentuates the enhancements in our approach. Our dataset offers clear advantages in terms of content variety, comment density, and overall usability.

\begin{figure}[htbp]
    \centering
    \scalebox{0.95}{
    \begin{tabular}{cccccc}
    \hline
    Dataset & Livebot & VideoIC & VideoLC & Twitch-FIFA  & LiveChat\\
    \hline
    \#Videos & 2,361 & 4,951 & 85 & 49 & 873\\
    \#Comments & 895,929 & 5,330,393 & 1,406,219 & 168,094 & 3,200,799\\
    Duration (h) & 114 & 557 & 175 & 86 & 438\\
    \hline
    Avg. Duration (s) & 174 & 405 & 7412 & 6318 & 1800\\
    Avg. \#Comments & 380 & 1,077 & 16,544 & 3430 & 3666\\
    Avg. \#Words & 5.42 & 5.39 & 6.53 & 5.55 & 4.02\\
    \hline
    Comment Density (c/s) & 2.18 & 2.66 & 2.23 & 0.54 & 2.02\\
    \hline
\end{tabular}}
    \caption{Comparative statistics of the datasets}
    \label{tab:1}
\end{figure}

By incorporating these features, our dataset provides a substantial contribution to the field, particularly for researchers and practitioners that focus on the English-speaking live video interaction domain. It thereby provides a unique and essential resource for those working on live comment generation in English-speaking platforms. A detailed comparison between the languages of the datasets and the collection websites is further illustrated in Table \ref{tab:2}.

\begin{figure}[htbp]
    \centering
    \begin{tabular}{c|ccccc}
    \hline
    Dataset & Livebot & VideoIC & VideoLC & Twitch-FIFA & LiveChat\\
    \hline
    Comments Type & Danmu & Danmu & Danmu & Live chat & Live chat\\
    Language & Chinese & Chinese & Chinese & English & English\\
    Website & Bilibili & Bilibili & QQ & Twitch & Twitch\\
    \hline
\end{tabular}
    \caption{Comparison of data in the different datasets}
    \label{tab:2}
\end{figure}

Viewers frequently employ a diverse array of emotes, which cannot be handled as conventional text. To accommodate this, we have gathered and stored all the emotes used in the collected comments, resulting in a collection of 44,716 different time-embedded emotes.

\begin{wrapfigure}{L}{0.57\textwidth}
  \centering
  \vspace{0.4cm}
  \includegraphics[scale=0.45]{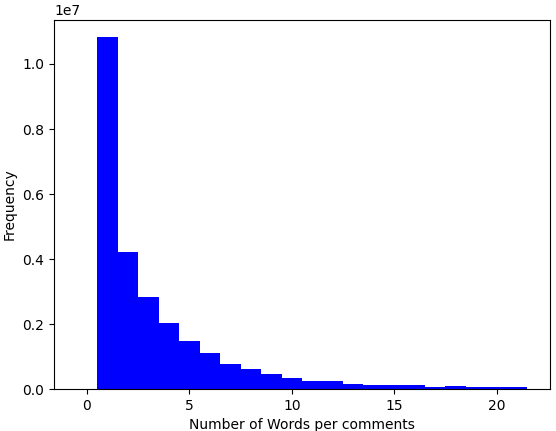}
  \caption{A histogram of the number of words per comments in the LiveChat dataset.}
  \label{fig:3}
\end{wrapfigure}
Owing to the real-time nature of the comments, it's not possible to guarantee the grammatical accuracy or the use of extended sentences by users. As depicted in Figure \ref{fig:3}, the histogram of the number of words per comment reveals that despite an average of 4.02 words per comment, the majority of them consist of just a single word. Furthermore, the diversity in the language used by the viewers contributes to a rich vocabulary size within the dataset, encompassing 541,811 unique words. This extensive vocabulary captures various expressions and reactions, which might pose additional challenges for model training but also offers a robust platform to evaluate the system's ability to understand and generate nuanced responses.
\section{Multimodal comments generation model}\label{transformer_model}

\begin{figure}[htbp]
  \centering
  \includegraphics[scale=0.13]{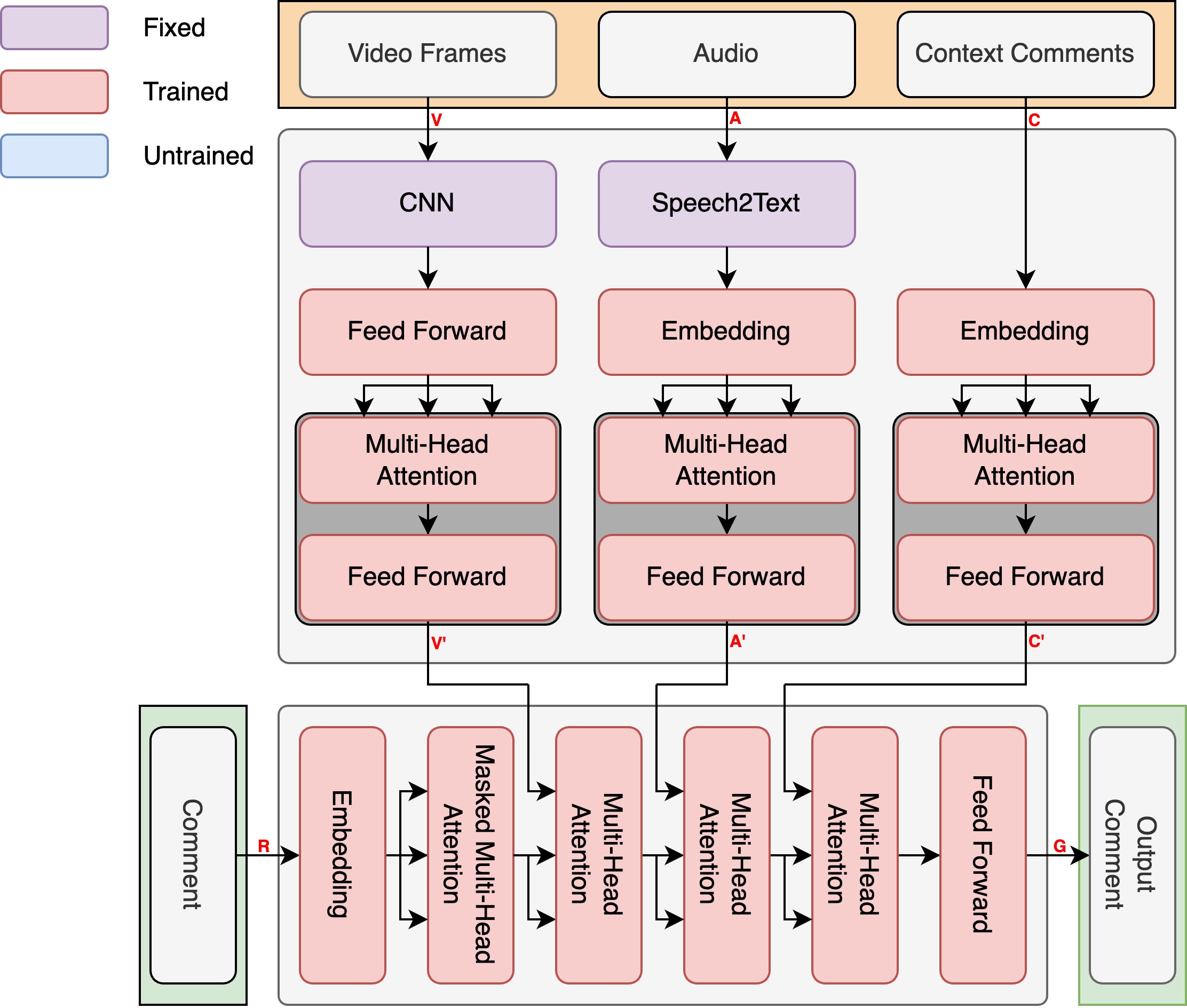}
  \caption{Architecture of the Triple Transformer Encoder Model.}
  \label{fig:2}
\end{figure}

We introduce our novel model designed for predicting live comments according to the Video, Audio and Comments contexts. The architecture efficiently integrates visual, audio, and textual information to produce relevant and coherent comments. Figure \ref{fig:2} illustrates the detailed structure of the proposed Triple Transformer Encoder model.

Figure \ref{tab:notation} shows a summary of the notations used below.

\subsection{Multi-Modal Encoder}
Our model follows an Encoder Decoder Architecture in which we first encode each modality into a latent representation which is then used to predict a relevant output comment. Every modality is encoded through a Multi Layer Transformers with $l_e$ layers and hidden sizes $d_v$ and $d_t$ for video and text inputs respectively.
\vspace{0.5cm}

\textbf{Video Encoder:} Given a timestep $t$, we sample one frame per second in the window $[t, t+T_1)$, $T_1$ being the length of the context window. We then obtain a vector $V=\{f_1, \dots, f_{T_1}\}$ of frames. Each frame is then encoded with a frozen ResNet50 to obtain the representation $V_F=\{v_1, \dots, v_{T_1}\} \in \mathbb{R}^{T_1\times d_v}$. We then add positional embedding to each frame before passing them through the Transformer Encoder. The final representation is then:
$$\begin{aligned}
  V'&=\{v'_1, \dots, v'_{T_1}\}\\
    &= Transformer_v(V_F)
    \in \mathbb{R}^{T_1\times d_v}
\end{aligned}$$

\textbf{Audio Encoder:} At a timestep $t$, we sample the audio from the window $[t, t+T_1)$, and pass it through the frozen Speech2Text model. This gives a representation $A=\{a\}=\{w^a_1, \dots, w^a_{p_a}\}$ which is given word and position embedding before being passed through the Audio Transformer Encoder producing the latent representation:
$$\begin{aligned}
  A'&=\{w^{'a}_1, \dots, w^{'a}_{p_a}\}\\
    &= Transformer_a(A)
    \in \mathbb{R}^{p_a\times d_t}
\end{aligned}$$

\textbf{Context Comments Encoder:} From the timestep $t$, we sample $n_c$ comments in the window $[t, t+T_1)$ to produce the vector $C=\{c_1, \dots, c_{n_c}\}\in \mathbb{R}^{p_a\times n_c}$. Each comment $c_i$ goes first through a word and positional embedding and is then encoded by the Comment Transformer Encoder to give the final representation of the comment $c^f_i=\{w^{'c}_1, \dots, w^{'c}_{n_c}\}$. We then denote $c'_i=w^{'c}_1$ which, we assume, contains the information encoded in the comment \textit{i}. Finally, we have a hidden representation of our context:
$$\begin{aligned}
  C'&=\{c'_1, \dots, c'_{n_c}\}\\
    &= \{Transformer_c(c_i), i\in [1:n_c]\}
    \in \mathbb{R}^{n_c\times d_t}
\end{aligned}$$

\subsection{Comment Decoder}

The decoder is responsible for taking the encoded representations from the previous section and using them to predict the target comment chosen in $[t+T_1, t+T_2)$, denoted by $R=\{r_1, \dots, r_{p_r}\}$. The decoder follows a modified Transformer architecture with three cross-attention layers, each designed to attend to a different modality: video, audio, and context comments.

\textbf{Target Comment Processing:} The target comment $R$ is, during training, the first input into the decoder. It goes through a word and positional embedding layer, followed by a masked self multi-head attention layer:

$$\begin{aligned}
R_{emb} &= Embedding(R)\\
R_{sa} &= SelfAttention(R_{emb})
\end{aligned}$$

The resulting processed target comment $R_{sa}$ serves as input for the cross-attention layers.

\textbf{Cross-Attention Layers:} The processed target comment is sequentially passed through three different cross-attention layers:
\begin{align*}
  R_v &= CrossAttention_v(V', R_{sa}, R_{sa}) \\
  R_{v, a} &= CrossAttention_a(A', R_v, R_v) \\
  R_{v, a, c} &= CrossAttention_c(C', R_{v, a}, R_{v, a})
       \in \mathbb{R}^{p_r \times d_t}
\end{align*}
Each cross-attention layer is tailored to attend to a specific modality, with $CrossAttention_v$ attending to the video, $CrossAttention_a$ attending to the audio, and $CrossAttention_c$ attending to the context comments.

\textbf{Prediction Layer:} After passing $R_{v, a, c}$ through a FeedForward layer, the prediction is done through a simple Linear layer mapping the final hidden state to the vocabulary.
$$\begin{aligned}
G &= Linear(FeedForward(R_{v, a, c}))
  \in \mathbb{R}^{p_r \times d_\Sigma}
\end{aligned}$$

\begin{figure}[htbp]
    \centering
    \scalebox{0.88}{
    \begin{tabular}{cc||cc}
    \hline
    Symbol & Definition & Symbol & Definition\\
    \hline
    $t$ & Starting timestep of the extract &$G$ & Generated Response vector\\
    $T_1$ & Length of the context window & $l_e$ & Layers in Transformer Decoder\\
    $T_2$ & Length of the clip & $l_d$ & Layers in Transformer Decoder\\
    $V$ & Video context vector & $d_v$ & Hidden size of the Video Transformer Encoder\\
    $A$ & Audio context vector & $d_t$ & Hidden size of the Text Transformers\\
    $C$ & Comments context vector& $p_a$ & Token count for audio text features\\
    $R$ & Response vector&$p_c$ & Token count for comments text features\\
    $V'$ & Latent video representation &$p_r$ & Token count for response text features\\
    $A'$ & Latent audio representation &$n_c$ & Number of context comments considered\\
    $C'$ & Latent comments representation&$d_\Sigma$ & Size of the Vocabulary\\

    \hline
\end{tabular}}
    \caption{Mathematical Notations}
    \label{tab:notation}
\end{figure}

\subsection{Training Setup}
The training process of our model consists of two primary stages: \textbf{Pretraining} and \textbf{Training}.

\textbf{Pretraining:} In this stage, the text encoders are pretrained using a masked language modeling (MLM) task. Given a target text \( Y = \{y_1, \dots, y_n\} \), a masked version \(\widetilde{Y} = \{\widetilde{y_1}, \dots, \widetilde{y_n}\}\) is created where \(\forall i, \widetilde{y_i} = [MASK]\) with probability \( p \) and \(\widetilde{y_i} = y_i\) with probability \( 1-p \). The model is then trained to predict the original tokens from the masked ones using the cross-entropy loss. This process enables the text encoders to capture the underlying semantics and structure of the language.
\begin{equation}
\mathcal{L}_{\text{pretrain}} = -\mathbb{E}_{y_i \sim Y, y_i \neq \widetilde{y_i}} \left[\log \left(p({y_i}|y_{\backslash i})\right) \right]
\end{equation}
\textbf{Training:} In the second stage, the entire model, including the video, audio, and context comment encoders, as well as the decoder, is trained. The objective in this phase is to predict the target comment from the multimodal input. The loss function is the cross-entropy loss, applied to the entire sequence of target tokens.

\begin{equation}
    \mathcal{L}_{\text{train}} = -\mathbb{E}_{y_i \sim Y} \left[\log \left(p({y_i}|y_{< i})\right) \right]
\end{equation}

In addition to the standard training setup, we employ a data augmentation technique to enhance the model's generalization capability. For every query, rather than having the model consistently predict a fixed target comment, we introduce randomness by selecting a comment uniformly at random from the response window. The model is then tasked with predicting this randomly chosen comment using the teacher-forcing technique. By not being bound to a single target for every query, the model is encouraged to generate a broader variety of comments, promoting richer and more diverse output during the generation phase.

\section{Experiment}
\subsection{Experiment setup}\label{setup}

We evaluate the performance of our model using various retrieval metrics. Given a query (in our case, the video, audio and comments contexts), and a set of candidates, retrieval metrics measure how well the model can retrieve or rank the relevant responses. The metrics we employ are:

\begin{itemize}
    \item \textbf{Recall@K:} Measures the percentage of times the true positive response is within the top \(K\) predictions. In our experiments, we compute Recall at 1, 2, and 5, denoted as Recall@1, Recall@2, and Recall@5 respectively.

    \item \textbf{Mean Rank (MR):} Average rank of the positive response among the list of candidate responses. Lower MR indicates better performance.

    \item \textbf{Mean Reciprocal Rank (MRR):} Average of the reciprocal ranks of the positive responses. For a particular query, if the rank of the true positive response is \( r \), the reciprocal rank is \( 1/r \). Higher MRR indicates better performance.
\end{itemize}

For each query, we select 10 candidate responses from the entire dataset. We adopt three different methodologies for this candidate selection:

\begin{enumerate}
    \item \textbf{Cosine Similarity:} Comments are chosen based on their cosine similarity with the chat context. Cosine similarity, which determines the cosine of the angle between two non-zero vectors, measures the semantic similarity in our context. By using this metric, we ensure that the selected candidates are closely related and contextually relevant to the chat sequence.

    \item \textbf{Popularity:} Comments are selected based on their frequency in a live stream, reflecting the most prevalent reactions or sentiments within the community. This approach captures the commonly repeated comments that resonate with a larger audience.

    \item \textbf{Random:} Candidates are picked at random from the dataset, providing an understanding of the baseline performance of the retrieval mechanism.
\end{enumerate}

By employing these methods, we aim to achieve a comprehensive view of the model's retrieval capabilities across varied selection criteria.\\

In the forthcoming iterations of our research, we plan to include a human evaluation segment. Human evaluators will assess the relevance and appropriateness of the model's responses to ensure that our automated metrics align with human perception and judgment.

\subsection{Results}\label{main_result}

In Table \ref{tab:results}, we present the performance metrics for our model under various candidate selection methods and data augmentation setups. These results have been obtained through meticulous experimentation and optimization of the model's hyperparameters. Specifically, for our optimal configuration, we set the learning rate to $10^{-4}$, batch size to $32$, pretrained the model for $100$ epochs and trained it for $200$ epochs. The Transformer encoders and decoders were configured with $4$ layers, $256$ hidden dimensions. We set the dropout to $0.1$ and choose $5$ and $15$ context comments during training and evaluation respectively. The table provides insights into how different selection methods and the application of data augmentation impact the model's retrieval capabilities.

\begin{figure}[htbp]
    \centering
    \scalebox{0.88}{
    \begin{tabular}{lcccccc}
\toprule
 & \multicolumn{3}{c}{Without Augmentation} & \multicolumn{3}{c}{With Augmentation} \\
\cmidrule(lr){2-4} \cmidrule(lr){5-7}
Method & Recall@1 & Recall@2 & Recall@5 & Recall@1 & Recall@2 & Recall@5 \\
\midrule
Cosine Similarity & 15.0 & 30.6 & 56.3 & 14.6 & 29.5 & 55.9 \\
Popularity & 18.9 & 32.6 & 59.4 & 18.3 & 32.0 & 58.6 \\
Randomness & 16.1 & 24.7 & 52.9 & 15.6 & 24.6 & 52.6 \\
\midrule
Method & Mean Rank & MRR & & Mean Rank & MRR & \\
\midrule
Cosine Similarity & 4.9 & 0.36 & & 5.0 & 0.35 & \\
Popularity & 4.7 & 0.37 & & 4.8 & 0.37 & \\
Randomness & 5.2 & 0.34 & & 5.2 & 0.33 & \\
\bottomrule
\end{tabular}}
    \caption{Evaluation results of the Triple Transformer Encoder Model on the LiveChat Dataset with and without the Data Augmentation technique}
    \label{tab:results}
\end{figure}

\section{Conclusion and Future Work}

In this research, we embarked on the intricate task of addressing live commenting in the streaming environment. Through scrupulous engineering and design, we proposed a Transformer-based model tailored to this context. Our work presented a large-scale, meticulously constructed dataset, encompassing an extensive range of comments, videos, categories, and streamers. This dataset stands as a core contribution of this study, devised to capture the rich, multifaceted nature of live-stream interactions.

The results indicate that our model holds its ground when juxtaposed with the Game-Based state-of-the-art benchmark in retrieval-based metrics. While our work sets an encouraging baseline, there is a discernible avenue for further enhancements.

Our approach currently does not fully utilize certain data aspects, including emotes and the timing within a live-stream, which represent potential for boosting the model's performance. Additionally, due to time constraints, a comprehensive human evaluation could not be undertaken. This leaves an element of uncertainty regarding how the model's outputs compare with genuine human comments.

Looking forward, refinements to the existing Transformer architecture are envisioned. This includes the integration of cross-modal interactions through mechanisms like Cross Attention, as well as embedding temporal information from the live-stream to generate contextually richer comments.

Furthermore, the exploration of alternative frameworks, such as CLIP [9], could provide intriguing insights. These architectures, although not a focus of this work, could offer innovative ways to synergize textual and visual information, possibly setting new benchmarks in performance.

In summation, this research marks a significant stride towards the development of intelligent live commenting systems, establishing a robust foundation for future endeavors in this burgeoning field.

\section{References}
[1] Jieting Chen, Junkai Ding, Wenping Chen, and Qin Jin. Knowledge enhanced model for live video comment generation. arXiv preprint arXiv:2304.14657, 2023.

[2] Junyoung Chung, Caglar Gulcehre, KyungHyun Cho, and Yoshua Bengio. Empirical evaluation of gated recurrent neural networks on sequence modeling. arXiv preprint arXiv:1412.3555, 2014.

[3] Jeffrey L Elman. Finding structure in time. Cognitive science, 14(2):179–211, 1990.

[4] Kaiming He, Xiangyu Zhang, Shaoqing Ren, and Jian Sun. Deep residual learning for image recognition. In Proceedings of the IEEE conference on computer vision and pattern recognition, pages 770–778, 2016.

[5] Sepp Hochreiter and Jürgen Schmidhuber. Long short-term memory. Neural computation, 9(8):1735–1780, 1997.

[6] Zhiheng Huang, Wei Xu, and Kai Yu. Bidirectional lstm-crf models for sequence tagging. arXiv preprint arXiv:1508.01991, 2015.

[7] Shuming Ma, Lei Cui, Damai Dai, Furu Wei, and Xu Sun. Livebot: Generating live video comments based on visual and textual contexts. In Proceedings of the AAAI Conference on Artificial Intelligence, volume 33, pages 6810–6817, 2019.

[8] Ramakanth Pasunuru and Mohit Bansal. Game-based video-context dialogue. arXiv preprint arXiv:1809.04560, 2018.

[9] Alec Radford, Jong Wook Kim, Chris Hallacy, Aditya Ramesh, Gabriel Goh, Sandhini Agarwal, Girish Sastry, Amanda Askell, Pamela Mishkin, Jack Clark, et al. Learning transferable visual models from natural language supervision. In International conference on machine learning, pages 8748–8763. PMLR, 2021.

[10] Alec Radford, Jong Wook Kim, Tao Xu, Greg Brockman, Christine McLeavey, and Ilya
Sutskever. Robust speech recognition via large-scale weak supervision. In International Conference on Machine Learning, pages 28492–28518. PMLR, 2023.

[11] Yuchen Ren, Yuan Yuan, and Lei Chen. Multi-modal guided attention for live video comments generation. In International Conference on Computer Graphics, Artificial Intelligence, and Data Processing (ICCAID 2021), volume 12168, pages 267–273. SPIE, 2022.

[12] Ashish Vaswani, Noam Shazeer, Niki Parmar, Jakob Uszkoreit, Llion Jones, Aidan N Gomez, Łukasz Kaiser, and Illia Polosukhin. Attention is all you need. Advances in neural information processing systems, 30, 2017.

[13] Weiying Wang, Jieting Chen, and Qin Jin. Videoic: A video interactive comments dataset and multimodal multitask learning for comments generation. In Proceedings of the 28th ACM International Conference on Multimedia, pages 2599–2607, 2020.
[14] Zhen Wang, Jianwen Zhang, Jianlin Feng, and Zheng Chen. Knowledge graph embedding by
translating on hyperplanes. 28(1), 2014.
[15] Hao Wu, Gareth JF Jones, and Francois Pitie. Response to livebot: Generating live video comments based on visual and textual contexts. arXiv preprint arXiv:2006.03022, 2020.

[16] Hao Wu, François Pitié, and Gareth Jones. Cold start problem for automated live video comments. In Proceedings of the Third Workshop on Multimodal Artificial Intelligence, pages54–62, 2021.

[17] Zehua Zeng, Neng Gao, Cong Xue, and Chenyang Tu. Plvcg: A pretraining based model for live video comment generation. In Pacific-Asia Conference on Knowledge Discovery and Data Mining, pages 690–702. Springer, 2021.

[18] Zhihan Zhang, Zhiyi Yin, Shuhuai Ren, Xinhang Li, and Shicheng Li. Dca: Diversified coattention towards informative live video commenting. In Natural Language Processing and Chinese Computing: 9th CCF International Conference, NLPCC 2020, Zhengzhou,China, October 14–18, 2020, Proceedings, Part II 9, pages 3–15. Springer, 2020.

\newpage
\bibliographystyle{plain}
\bibliography{bibfile.bib}

\end{document}